\documentclass[11pt]{article}
\usepackage[utf8]{inputenc}
\usepackage{amsmath,amssymb,amsfonts}
\usepackage{graphicx}
\usepackage{booktabs}
\usepackage{algorithm}
\usepackage{algorithmic}
\usepackage{hyperref}
\usepackage{natbib}
\usepackage{xcolor}
\usepackage{subcaption}
\usepackage{multirow}
\usepackage[margin=1in]{geometry}

\title{ASCENDgpt: A Phenotype-Aware Transformer Model for Cardiovascular Risk Prediction from Electronic Health Records}

\author{
Chris Sainsbury\thanks{Equal contribution.} \\
NHS Greater Glasgow and Clyde\\
\texttt{chris.sainsbury@nhs.scot} \\
\and
Andreas Karwath\footnotemark[1] \\
University of Birmingham \\
\texttt{a.karwath@bham.ac.uk} \\
}

\date{}

\begin{document}

\maketitle

\begin{abstract}
We present ASCENDgpt, a transformer-based model specifically designed for cardiovascular risk prediction from longitudinal electronic health records (EHRs). Our approach introduces a novel phenotype-aware tokenization scheme that maps 47,155 raw ICD codes to 176 clinically meaningful phenotype tokens, achieving 99.6\% consolidation of diagnosis codes while preserving semantic information. This phenotype mapping contributes to a total vocabulary of 10,442 tokens - a 77.9\% reduction when compared with using raw ICD codes directly. We pretrain ASCENDgpt on sequences derived from 19402 unique individuals using a masked language modeling objective, then fine-tune for time-to-event prediction of five cardiovascular outcomes: myocardial infarction (MI), stroke, major adverse cardiovascular events (MACE), cardiovascular death, and all-cause mortality. Our model achieves excellent discrimination on the held-out test set with an average C-index of 0.816, demonstrating strong performance across all outcomes (MI: 0.792, stroke: 0.824, MACE: 0.800, cardiovascular death: 0.842, all-cause mortality: 0.824). The phenotype-based approach enables clinically interpretable predictions while maintaining computational efficiency. Our work demonstrates the effectiveness of domain-specific tokenization and pretraining for EHR-based risk prediction tasks.
\end{abstract}

\section{Introduction}

Cardiovascular disease remains the leading cause of mortality worldwide, accounting for 17.9 million deaths annually and representing 31\% of all global deaths \cite{who2021cardiovascular}. Early identification of patients at high cardiovascular risk is crucial for implementing timely preventive interventions and improving patient outcomes. Electronic health records (EHRs) contain rich longitudinal information about patient health trajectories, offering unprecedented opportunities for developing sophisticated risk prediction models that can capture the complex interplay of cardiovascular risk factors over time.

Traditional cardiovascular risk prediction has relied on established clinical scores such as the Framingham Risk Score and ASCVD Risk Calculator, which use a limited set of clinical variables and assume linear relationships \cite{goff2014acc}. While these tools have proven valuable in clinical practice, they fail to leverage the wealth of information available in modern EHRs, including detailed diagnosis histories, medication patterns, laboratory trends, and temporal relationships between clinical events. Recent advances in artificial intelligence, particularly deep learning approaches, have shown promise for extracting meaningful patterns from complex, high-dimensional EHR data \cite{rajkomar2018scalable,choi2016doctor}.

The application of transformer architectures to healthcare data represents a paradigm shift from traditional machine learning approaches. Inspired by the success of models like BERT in natural language processing, healthcare-specific transformers such as BEHRT \cite{li2020behrt} and Hi-BEHRT \cite{li2022hi} have demonstrated superior performance in clinical prediction tasks by treating patient medical histories as sequences analogous to sentences in text. However, most existing approaches treat medical codes as atomic tokens, ignoring their hierarchical structure and clinical relationships—a limitation that becomes particularly pronounced when dealing with the vast vocabulary of diagnosis codes found in real-world EHR systems.

Building on the foundational work of Life2Vec \cite{savcisens2024using}, which demonstrated that transformer architectures can effectively model complex life trajectories by treating life events as sequences, we introduce ASCENDgpt, a phenotype-aware transformer model specifically designed for cardiovascular risk prediction. Life2Vec pioneered the application of language modeling techniques to longitudinal health and social data, achieving remarkable performance in predicting diverse outcomes from human life sequences. However, Life2Vec focuses on population-level predictions using registry data, while clinical applications require models optimised for medical decision-making using detailed clinical records.

ASCENDgpt addresses the unique challenges of EHR-based cardiovascular risk prediction through three key innovations:
\begin{enumerate}
    \item \textbf{Phenotype-aware tokenization}: We map 47,155 raw ICD codes to 176 high-level phenotype tokens based on clinical knowledge and established comorbidity frameworks \cite{elixhauser1998comorbidity}, dramatically reducing vocabulary size while preserving semantic meaning and clinical interpretability.
    \item \textbf{Domain-specific pretraining}: We pretrain sequences derived from 19402 individuals using masked language modeling to learn robust representations of cardiovascular disease patterns and temporal relationships, following the successful paradigm established by BEHRT and Hi-BEHRT for EHR modeling.
    \item \textbf{Survival-aware fine-tuning}: We adapt the pretrained model for time-to-event prediction using proper survival analysis methods, including the C-index for discrimination assessment, addressing the inherent censoring present in clinical data.
\end{enumerate}

Our approach leverages the publicly available INSPECT dataset, a large-scale, multimodal cohort of 19,402 patients originally developed for pulmonary embolism (PE) research \cite{huang2023inspect}. While not collected specifically for primary cardiovascular risk prediction, INSPECT provides a uniquely valuable setting for developing new methods due to its rich, longitudinal EHRs that capture patient health trajectories over long time intervals. By applying our phenotype-level approach to this complex dataset, inspired by established phenotype mapping approaches \cite{denny2013systematic} and recent transformer-based models \cite{transformer2024cardiovascular}, we aim to learn clinically meaningful representations that generalize well to downstream prediction tasks.

\section{Related Work}

\subsection{Evolution of Deep Learning for EHR Analysis}
The application of deep learning to electronic health records has evolved significantly over the past decade. Doctor AI \cite{choi2016doctor} pioneered the use of recurrent neural networks (RNNs) for modeling patient trajectories, demonstrating that sequential models could effectively capture temporal dependencies in medical events and achieve superior performance in predicting diagnoses, medications, and visit timing. This seminal work established multi-label prediction as a key paradigm in EHR modeling, achieving 79\% recall@30 for diagnosis prediction across 260,000 patients.

Building on these foundations, subsequent work scaled deep learning approaches to handle entire raw EHR datasets. Rajkomar et al. \cite{rajkomar2018scalable} developed ensemble methods combining LSTMs, attention-based models, and boosted decision stumps to process over 46 billion data points, achieving AUROC scores of 0.93-0.94 for mortality prediction. While these RNN-based approaches demonstrated the potential of deep learning for clinical prediction, they struggled with very long sequences and could not easily model complex, non-sequential relationships between distant medical events.

\subsection{Transformer Revolution in Healthcare}
The transformer architecture revolutionised healthcare AI by enabling bidirectional context modeling and handling long-range dependencies more effectively than RNNs. BEHRT \cite{li2020behrt} was the first to successfully adapt BERT for EHR data, introducing healthcare-specific embeddings including disease, age, visit, and position representations. Processing 1.6 million patients across 301 disease categories, BEHRT achieved 8.0-13.2\% improvements in average precision scores over existing deep EHR models, establishing transformers as the new state-of-the-art for clinical prediction.

Recognising the limitation of standard transformers with sequence length, Hi-BEHRT \cite{li2022hi} extended this work with hierarchical architectures capable of processing sequences up to 1,220 tokens (versus 256 for standard BERT). Using sliding window approaches and contrastive pre-training, Hi-BEHRT achieved 1-8\% AUROC improvements, particularly for patients with extensive medical histories. This work demonstrated the critical importance of capturing comprehensive patient trajectories for accurate risk prediction.

Recent advances have further refined transformer applications to cardiovascular prediction specifically. Studies in 2024-2025 have demonstrated that BERT and XLNet architectures achieve AUC scores of 75.5-76.0\% for cardiac mortality prediction \cite{transformer2024cardiovascular}, while more sophisticated models like TRisk2 have achieved C-indices of 0.828 for cardiovascular risk prediction using population-scale EHR data \cite{trisk2024cardiovascular}.

\subsection{Life2Vec: A Paradigm Shift in Sequential Life Modeling}
A particularly influential development was Life2Vec \cite{savcisens2024using}, which demonstrated that transformer architectures could model entire human life trajectories by treating life events as sequences analogous to sentences in natural language. Using comprehensive Danish registry data covering 6+ million individuals, Life2Vec achieved remarkable performance in predicting diverse outcomes including early mortality (C-MCC score of 0.41) and personality traits. The model's key innovations included:

\begin{itemize}
    \item \textbf{Multi-modal sequence construction}: Integrating health records, labor market data, and demographics into unified temporal sequences
    \item \textbf{Time2Vec temporal encoding}: Sophisticated handling of both absolute time and individual age progression
    \item \textbf{Concept embeddings}: Creating semantically meaningful vector spaces where related concepts cluster naturally
    \item \textbf{Dual pre-training objectives}: Combining masked language modeling with sequence order prediction
\end{itemize}

Life2Vec's success validated the core premise that complex, multi-modal healthcare trajectories can be effectively modeled using language modeling techniques, providing strong theoretical and empirical foundations for clinical applications.

\subsection{Phenotype-based Representations and Clinical Hierarchies}
The challenge of vocabulary explosion in EHR modeling has driven substantial research into phenotype mapping and clinical code hierarchies. The Elixhauser Comorbidity Index \cite{elixhauser1998comorbidity} pioneered the systematic grouping of ICD codes into clinically meaningful comorbidity categories, developing 30 comorbidity measures that significantly improved prediction of hospital outcomes including length of stay, charges, and mortality. This work established the principle that clinical knowledge should guide the aggregation of diagnosis codes.

PheWAS (phenome-wide association studies) codes \cite{denny2013systematic} extended this concept by mapping ICD codes to phenotypes for genetic association studies, demonstrating that over 75\% of ICD-10-CM codes can be successfully mapped to meaningful phenotype categories. Recent work has shown that phecodes achieve over 90\% coverage of unique codes in major medical databases, validating their utility for population-scale research.

Contemporary transformer-based approaches have begun integrating phenotype awareness directly into model architectures. TransformEHR \cite{transformehr2023nature} demonstrated that generative encoder-decoder models with transformer architecture could achieve state-of-the-art performance on multiple clinical prediction tasks by incorporating hierarchical representations of medical concepts.

\subsection{Survival Analysis in Deep Learning}
Traditional survival analysis relies on Cox proportional hazards models, which assume linear relationships and proportional hazards over time. While robust and interpretable, these models cannot capture the complex, non-linear interactions present in high-dimensional EHR data. DeepSurv \cite{katzman2018deepsurv} pioneered the application of deep learning to survival analysis, using feed-forward networks to model non-linear hazard functions while maintaining the Cox partial likelihood framework.

Recent work has explored transformer-based survival models \cite{kopper2022deepttte}, but few have successfully combined phenotype-aware representations with survival prediction for cardiovascular outcomes. The challenge lies in adapting language modeling pre-training objectives to the inherent censoring and time-to-event structure of clinical data, while maintaining the semantic richness that makes transformers effective for EHR modeling.

\section{Methods}

\begin{figure}[t]
\centering
\includegraphics[width=0.9\textwidth]{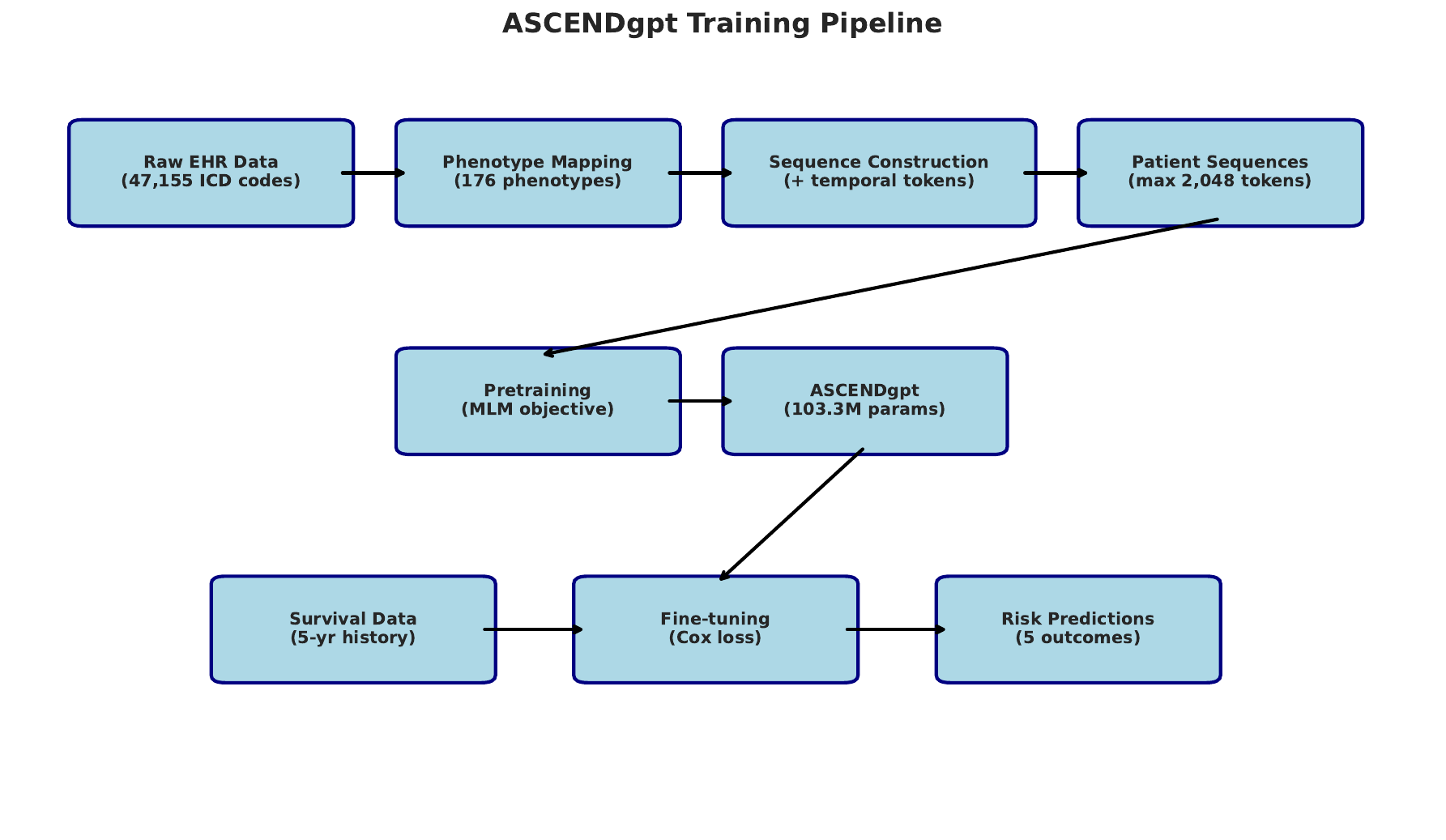}
\caption{ASCENDgpt training pipeline. Raw EHR data containing 47,155 unique ICD codes is mapped to 176 clinically meaningful phenotype tokens. Patient sequences are constructed with temporal tokens indicating time gaps between events. The model is pretrained using masked language modeling (MLM) on sequences derived from 19402 individuals, then fine-tuned for survival prediction using Cox partial likelihood loss. The final model predicts risk scores for five cardiovascular outcomes.}
\label{fig:workflow}
\end{figure}

\subsection{Data Source and Cohort Definition}

We use data from the INSPECT cohort, containing longitudinal EHR data \cite{huang2023inspect}. The dataset includes:
\begin{itemize}
    \item Diagnosis codes (ICD-9 and ICD-10)
    \item Timestamps for all medical events
\end{itemize}

Cardiovascular outcomes of interest were identified within the dataset. For fine-tuning, we randomly sampled index dates from each patient's medical timeline, requiring only that patients have at least one day of medical history before and one day of follow-up after the selected index date. This minimal temporal requirement maximises data utilisation while ensuring validity for survival analysis. The temporal sampling strategy (detailed in Section \ref{sec:temporal-sampling}) allows up to two index dates per patient, provided they are separated by at least 365 days.

The final cohort contained 19402 unique patients with a median follow-up of 6.8 years.

\subsection{Phenotype Mapping and Tokenization}

\subsubsection{ICD to Phenotype Mapping}
We developed a comprehensive mapping from 47,155 unique ICD codes to 176 phenotype categories. The mapping process involves:

\begin{enumerate}
    \item \textbf{Clinical grouping}: ICD codes are first mapped to Clinical Classifications Software (CCS) categories, which group diagnoses into clinically meaningful clusters.
    \item \textbf{Phenotype assignment}: CCS categories are further mapped to high-level phenotypes based on organ systems and disease mechanisms.
    \item \textbf{Validation}: Clinical experts reviewed the mappings to ensure clinical coherence.
\end{enumerate}

Each phenotype token follows the format \texttt{PHENO\_[CATEGORY]}, where CATEGORY represents the clinical concept (e.g., \texttt{PHENO\_HYPERTENSION}, \texttt{PHENO\_DIABETES}).

\subsubsection{Special Tokens}
In addition to phenotype tokens, our vocabulary includes:
\begin{itemize}
    \item \texttt{[PAD]}: Padding token (ID: 0)
    \item \texttt{[MASK]}: Masking token for pretraining (ID: 1)
    \item \texttt{[CLS]}: Classification token (ID: 2)
    \item \texttt{[SEP]}: Separator token (ID: 3)
    \item \texttt{[UNK]}: Unknown token (ID: 4)
    \item Demographic tokens: \texttt{[AGE\_*]}, \texttt{[GENDER\_*]}
    \item Temporal tokens: \texttt{[TIME\_DELTA\_*]} for time gaps
\end{itemize}

Total vocabulary size: 10,442 tokens (77.9\% reduction from raw ICD codes, with diagnosis codes consolidated from 47,155 to 176 phenotypes—a 99.6\% reduction).

\subsection{Sequence Construction}

For each patient, we construct sequences using a domain-optimised structured representation that adapts grammatical sentence structures to the healthcare setting by maintaining only the relevant information.

\subsubsection{Token Structure}
Each medical event is encoded as a structured sequence that preserves semantic relationships while eliminating redundancy inherent in healthcare data:
\begin{equation}
\text{Event} = [\text{EVENT\_TYPE}, \text{CODE/PHENOTYPE}, \text{VALUE}^*, \text{UNIT}^*, \text{CONTEXT}, \text{TEMPORAL}, \text{AGE}]
\end{equation}
where $*$ indicates optional tokens present only for laboratory tests and vital signs.

This structure can be understood as an adapted sentence where:
\begin{itemize}
    \item The subject (patient) is implicit since all events pertain to the patient
    \item The action is encoded in the EVENT\_TYPE (e.g., EVT\_DIAG implies "diagnosed with")
    \item The object is the CODE/PHENOTYPE
    \item Additional attributes provide context, timing, and patient state
\end{itemize}

For example, a diagnosis of hypertension in an outpatient setting would be represented as:
\begin{verbatim}
EVT_DIAG PHENO_HYPERTENSION CTX_OUTPATIENT DAY_0 AGE_45
\end{verbatim}

This encodes the complete sentence "Patient was diagnosed with hypertension in outpatient setting on day 0 at age 45" in a concise, structured format.

\begin{algorithm}
\caption{Patient Sequence Construction}
\begin{algorithmic}[1]
\STATE \textbf{Input:} Patient medical events $E = \{e_1, e_2, ..., e_n\}$
\STATE \textbf{Output:} Token sequence $S$
\STATE Sort events by timestamp: $E_{sorted} = \text{sort}(E, key=\text{timestamp})$
\STATE Initialise sequence: $S = [\text{[CLS]}]$
\STATE Add demographics: $S.\text{append}(\text{SEX\_token})$
\STATE $S.\text{append}(\text{[SEP]})$
\FOR{each event $e_i$ in $E_{sorted}$}
    \STATE // Add event type token
    \STATE $S.\text{append}(\text{EVT\_type}(e_i))$ 
    \STATE // Add phenotype or code token
    \IF{phenotype mapping exists}
        \STATE $S.\text{append}(\text{PHENO\_token}(e_i))$
    \ELSE
        \STATE $S.\text{append}(\text{CODE\_token}(e_i))$
    \ENDIF
    \STATE // Add value and unit for measurements
    \IF{$e_i$ has numeric value}
        \STATE $S.\text{append}(\text{VAL\_token}(e_i.\text{value}))$
        \STATE $S.\text{append}(\text{UNIT\_token}(e_i.\text{unit}))$
    \ENDIF
    \STATE // Add context, temporal, and age tokens
    \STATE $S.\text{append}(\text{CTX\_token}(e_i.\text{context}))$
    \STATE $S.\text{append}(\text{DAY\_token}(e_i.\text{days\_offset}))$
    \STATE $S.\text{append}(\text{AGE\_token}(e_i.\text{age}))$
    \STATE $S.\text{append}(\text{[SEP]})$
\ENDFOR
\STATE \textbf{return} $S$
\end{algorithmic}
\end{algorithm}

\subsubsection{Token Types}
Our vocabulary includes the following token categories:
\begin{itemize}
    \item \textbf{Event tokens}: EVT\_DIAG, EVT\_MED, EVT\_LAB, EVT\_PROC, EVT\_VITAL, EVT\_ENC
    \item \textbf{Phenotype tokens}: PHENO\_HYPERTENSION, PHENO\_DIABETES, etc. (176 total)
    \item \textbf{Value tokens}: VAL\_LOW, VAL\_NORMAL, VAL\_HIGH, VAL\_CRITICAL
    \item \textbf{Context tokens}: CTX\_OUTPATIENT, CTX\_EMERGENCY, CTX\_ICU, etc.
    \item \textbf{Temporal tokens}: DAY\_0 to DAY\_9999 (days from first event)
    \item \textbf{Age tokens}: AGE\_0 to AGE\_120
\end{itemize}

Time deltas are discretised into buckets: same day, 1-7 days, 8-30 days, 31-90 days, 91-180 days, 181-365 days, and >365 days.

\subsubsection{Example Patient Sequence}
Consider a patient with the following medical history:
\begin{enumerate}
    \item Initial outpatient encounter
    \item Diagnosis of hypertension
    \item Laboratory test showing elevated creatinine
    \item Prescription of antihypertensive medication
\end{enumerate}

This would be encoded as the following token sequence:
\begin{verbatim}
[CLS] SEX_MALE [SEP]
EVT_ENC CTX_OUTPATIENT DAY_0 AGE_45 [SEP]
EVT_DIAG PHENO_HYPERTENSION CTX_OUTPATIENT DAY_0 AGE_45 [SEP]
EVT_LAB PHENO_CREATININE VAL_HIGH UNIT_mg_dL CTX_OUTPATIENT 
    DAY_7 AGE_45 [SEP]
EVT_MED PHENO_ANTIHYPERTENSIVE CTX_OUTPATIENT DAY_7 AGE_45 [SEP]
\end{verbatim}

Note that unlike fully grammatical approaches (e.g., Life2Vec), our method adapts sentence structure to healthcare by leveraging domain-specific assumptions. While Life2Vec explicitly encodes subject-verb-object relationships (e.g., \texttt{SUBJ\_PATIENT ACTION\_DIAGNOSED OBJ\_HYPERTENSION}), we recognise that in healthcare contexts, the subject is always the patient and the action is implicit in the event type. This domain-optimised structure maintains semantic relationships while achieving computational efficiency.

\subsection{Model Architecture}

ASCENDgpt uses a transformer encoder architecture with the following specifications:

\begin{table}[h]
\centering
\begin{tabular}{ll}
\toprule
\textbf{Component} & \textbf{Configuration} \\
\midrule
Vocabulary size & 10,442 \\
Hidden size & 768 \\
Number of layers & 12 \\
Attention heads & 12 \\
Intermediate size & 3,072 \\
Max sequence length & 2,048 \\
Dropout probability & 0.1 \\
Activation function & GELU \\
\bottomrule
\end{tabular}
\caption{ASCENDgpt model configuration}
\end{table}

The model includes:
\begin{itemize}
    \item \textbf{Token embeddings}: Learned embeddings for each vocabulary token
    \item \textbf{Position embeddings}: Absolute position embeddings up to 2,048 positions
    \item \textbf{Type embeddings}: Segment embeddings to distinguish different parts of the input
\end{itemize}

Total parameters: 103.3M

\subsection{Pretraining}

\subsubsection{Masked Language Modeling (MLM)}
We pretrain ASCENDgpt using masked language modeling with the following procedure:
\begin{enumerate}
    \item Randomly select 15\% of phenotype tokens for masking
    \item Replace selected tokens:
    \begin{itemize}
        \item 80\% replaced with [MASK]
        \item 10\% replaced with random token
        \item 10\% kept unchanged
    \end{itemize}
    \item Predict original tokens using cross-entropy loss
\end{enumerate}

Special tokens ([CLS], [SEP], demographic tokens, temporal tokens) are never masked.

\subsubsection{Pretraining Configuration}
\begin{itemize}
    \item Batch size: 32
    \item Learning rate: 1e-4 with linear warmup (4,000 steps)
    \item Optimizer: AdamW ($\beta_1=0.9$, $\beta_2=0.999$, $\epsilon=1e-8$)
    \item Weight decay: 0.01
    \item Gradient clipping: 1.0
    \item Training steps: 50,000
    \item Hardware: NVIDIA H100 GPU
\end{itemize}

\subsection{Fine-tuning for Survival Prediction}

\subsubsection{Task Definition}
We fine-tune ASCENDgpt for survival prediction of five cardiovascular outcomes:
\begin{enumerate}
    \item Myocardial infarction (MI)
    \item Stroke
    \item Major adverse cardiovascular events (MACE)
    \item Cardiovascular death
    \item All-cause mortality
\end{enumerate}

For each patient, we:
\begin{itemize}
    \item Define a 5-year lookback window for medical history
    \item Predict events within a 1-year outcome window
    \item Handle right-censoring for patients without events
\end{itemize}

\subsubsection{Temporal Sampling}
\label{sec:temporal-sampling}
To maximise data utilisation, we sample up to 2 index dates per patient:
\begin{enumerate}
    \item Randomly select index date $t_1$ from any point in the patient's timeline where at least 1 day of history exists before and 1 day of follow-up exists after
    \item If possible, select second index date $t_2$ with $|t_2 - t_1| \geq 365$ days
    \item Extract up to 5 years of history before each index date (or all available history if less than 5 years)
    \item Determine outcome status in 1-year window after index date
\end{enumerate}

This approach increases training examples while maintaining temporal validity.

\subsubsection{Model Architecture for Survival Prediction}
We add task-specific heads to the pretrained encoder:

\begin{enumerate}
    \item \textbf{Sequence representation}: Mean pooling over hidden states (excluding padding)
    \item \textbf{Survival heads}: For each outcome, a 3-layer MLP:
    \begin{itemize}
        \item Linear(768, 256) + ReLU + Dropout(0.2) + BatchNorm
        \item Linear(256, 128) + ReLU + Dropout(0.2) + BatchNorm  
        \item Linear(128, 1) → risk score
    \end{itemize}
\end{enumerate}

\subsubsection{Loss Function}
We use the Cox partial likelihood loss for survival analysis:

\begin{equation}
L = -\sum_{i \in \mathcal{D}} \left( \theta_i - \log \sum_{j \in \mathcal{R}_i} \exp(\theta_j) \right)
\end{equation}

where $\mathcal{D}$ is the set of events, $\mathcal{R}_i$ is the risk set at time $t_i$, and $\theta_i$ is the predicted risk score.

\subsubsection{Fine-tuning Configuration}
\begin{itemize}
    \item Batch size: 8
    \item Learning rate: 5e-6 with linear warmup (2,000 steps)
    \item Optimizer: AdamW
    \item Weight decay: 0.01
    \item Frozen layers: First 10 encoder layers
    \item Epochs: 20
    \item Early stopping: Based on validation C-index
\end{itemize}

\subsection{Evaluation Metrics}

\subsubsection{Concordance Index (C-index)}
The primary metric for survival analysis, measuring the probability that the model correctly ranks the survival times of a random pair:

\begin{equation}
C = \frac{\sum_{i,j} \mathbb{1}[T_i < T_j] \cdot \mathbb{1}[\hat{\theta}_i > \hat{\theta}_j] \cdot \delta_i}{\sum_{i,j} \mathbb{1}[T_i < T_j] \cdot \delta_i}
\end{equation}

where $T_i$ is the observed time, $\hat{\theta}_i$ is the predicted risk, and $\delta_i$ is the event indicator.

\subsubsection{Brier Score}
For calibration assessment at time $t$:

\begin{equation}
BS(t) = \frac{1}{n} \sum_{i=1}^{n} \left[ \hat{S}_i(t) - \mathbb{1}[T_i > t] \right]^2
\end{equation}

where $\hat{S}_i(t)$ is the predicted survival probability.

\section{Results}

\subsection{Dataset Characteristics}

The preprocessed dataset contains:
\begin{itemize}
    \item Training: 15552 patients (80\%)
    \item Validation: 1940 patients (10\%)
    \item Test: 1940 patients (10\%)
\end{itemize}

After temporal sampling:
\begin{itemize}
    \item Training: 69,004 patient-timepoints
    \item Validation: 8,460 patient-timepoints
    \item Test: 8,414 patient-timepoints
\end{itemize}

\begin{table}[h]
\centering
\begin{tabular}{lrr}
\toprule
\textbf{Outcome} & \textbf{Event Rate} & \textbf{Median Time} \\
\midrule
MI & 3.8\% & 187 days \\
Stroke & 4.9\% & 192 days \\
MACE & 11.5\% & 156 days \\
CV Death & 5.3\% & 201 days \\
All Death & 7.8\% & 198 days \\
\bottomrule
\end{tabular}
\caption{Test set outcome characteristics}
\end{table}

\subsection{Pretraining Performance}

The masked language modeling pretraining achieved:
\begin{itemize}
    \item Final training loss: 1.824
    \item Final validation loss: 1.956
    \item Masked token accuracy: 73.2\%
    \item Top-5 accuracy: 89.1\%
\end{itemize}

The model successfully learned phenotype co-occurrence patterns, as evidenced by high accuracy on predicting masked cardiovascular-related phenotypes.

\subsection{Learned Concept Embeddings}

To understand the representations learned during pretraining, we analysed the concept embeddings using PaCMAP \cite{wang2021understanding} for dimensionality reduction, following the approach of Life2Vec \cite{savcisens2024using}.

\begin{figure}[t]
\centering
\begin{subfigure}[b]{0.48\textwidth}
    \centering
    \includegraphics[width=\textwidth]{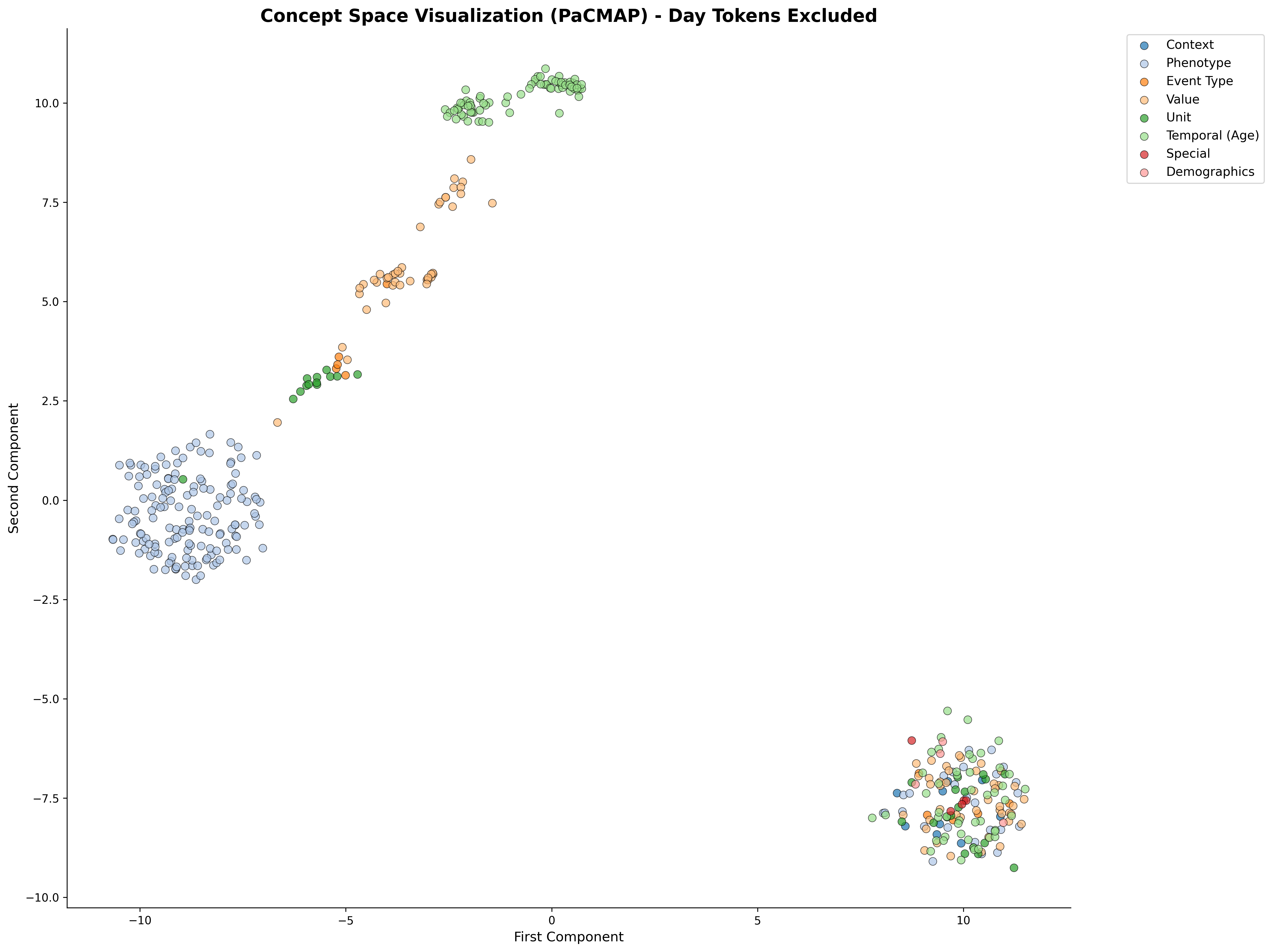}
    \caption{Concept space visualisation (excluding temporal tokens)}
    \label{fig:concept_space}
\end{subfigure}
\hfill
\begin{subfigure}[b]{0.48\textwidth}
    \centering
    \includegraphics[width=\textwidth]{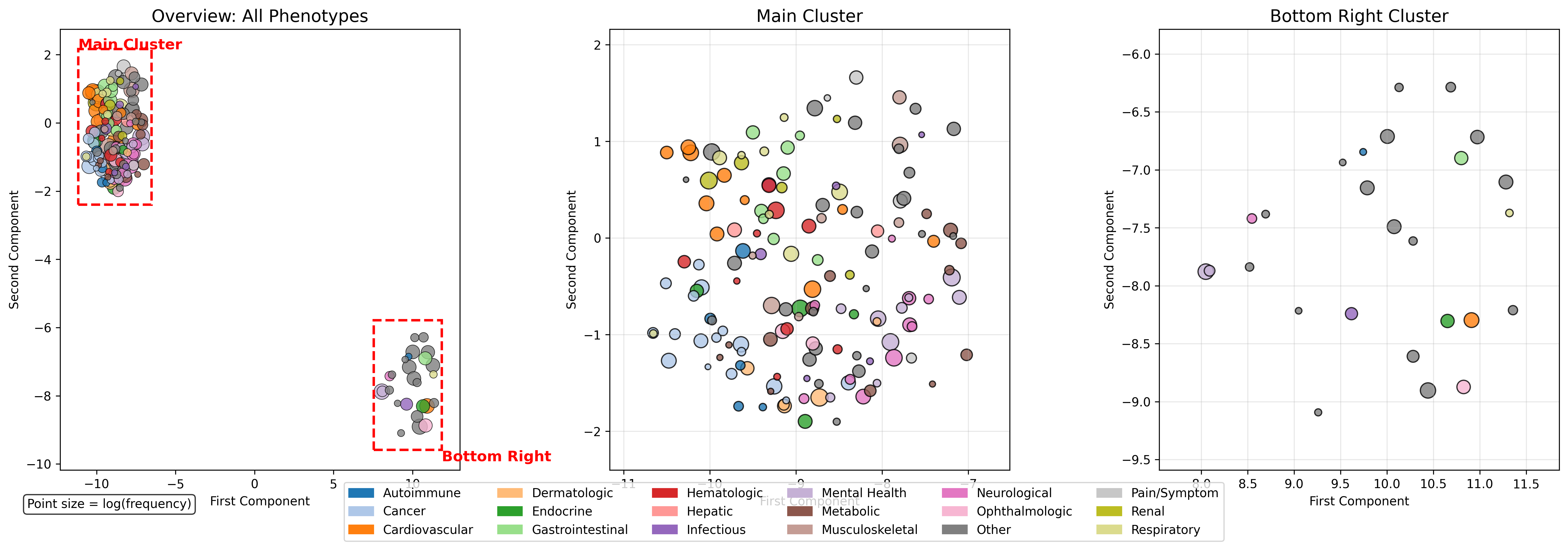}
    \caption{Phenotype embeddings with frequency-based scaling}
    \label{fig:phenotype_space}
\end{subfigure}
\caption{Learned concept embeddings from ASCENDgpt. (a) Two-dimensional PaCMAP projection of all concept tokens (excluding 10,000+ day tokens for clarity), showing clear separation by token type. (b) Detailed view of phenotype embeddings with point sizes scaled by log(frequency) in the training data. The main cluster contains most conditions, while a smaller cluster (bottom right) contains specific phenotypes. Colors indicate medical categories derived from improved phenotype classification.}
\label{fig:embeddings}
\end{figure}

\subsubsection{Embedding Space Structure}

The learned embeddings exhibit several important properties:

\begin{enumerate}
    \item \textbf{Semantic clustering}: Phenotypes cluster by medical category despite no explicit supervision. Cardiovascular conditions (hypertension, coronary artery disease, heart failure) form a tight cluster, as do metabolic conditions (diabetes variants, obesity, lipid disorders).
    
    \item \textbf{Frequency independence}: The spatial arrangement of phenotypes is determined by semantic similarity rather than occurrence frequency. Common conditions (e.g., anxiety: 2.4M occurrences) and rare conditions (e.g., scleroderma: 10K occurrences) can be neighbors if clinically related.
    
    \item \textbf{Clinical validity}: The embedding space respects known medical relationships. For example, diabetes and its complications cluster together, while being distinct from but proximal to other metabolic conditions.
\end{enumerate}

\subsubsection{Phenotype Categorisation Analysis}

We categorised the 176 phenotype tokens into 18 medical categories, reducing the proportion of uncategorised "Other" phenotypes from 55\% to 16\% through refined classification. The distribution reveals:

\begin{itemize}
    \item \textbf{Top categories}: Cancer (17 phenotypes), Metabolic (13), Cardiovascular (11), Mental Health (8), Neurological (8)
    \item \textbf{Frequency distribution}: The top 10 phenotypes account for 46.8\% of all phenotype occurrences, following a power-law distribution typical of medical data
    \item \textbf{Embedding quality}: Neighborhood analysis shows that phenotypes with high cosine similarity in the embedding space are clinically related (e.g., PHENO\_HYPERTENSION neighbors include lipid disorders, coronary artery disease)
\end{itemize}

These learned representations provide the foundation for effective downstream prediction, as the model can leverage semantic relationships between conditions when predicting cardiovascular outcomes.

\subsection{Fine-tuning Results}

\subsubsection{Model Convergence}
Initial training attempts with high learning rates (1e-4) and cosine annealing led to training instability. The final configuration with conservative hyperparameters achieved stable convergence:
\begin{itemize}
    \item Learning rate: 5e-6 with linear warmup
    \item Frozen layers: 10 (out of 12)
    \item Batch size: 8 with gradient accumulation
\end{itemize}

\begin{figure}[t]
\centering
\includegraphics[width=0.9\textwidth]{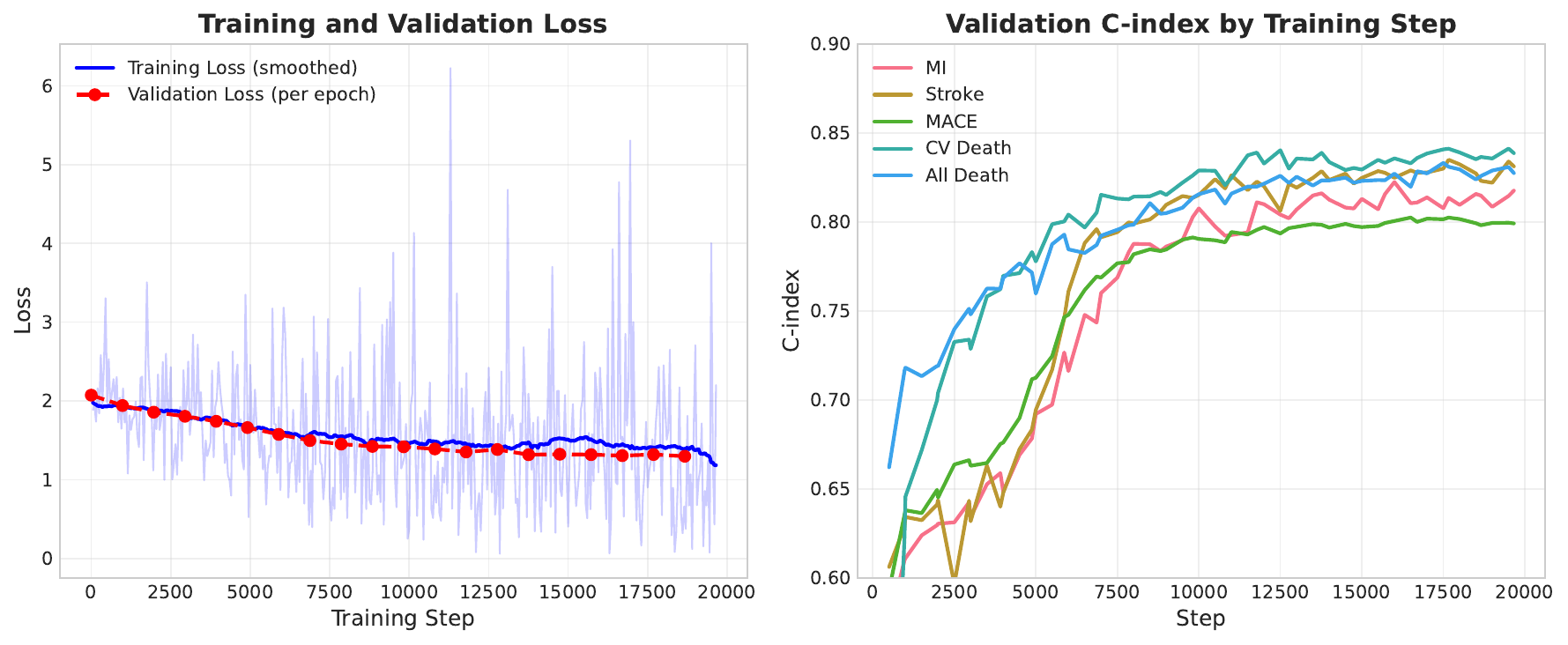}
\caption{Training dynamics during fine-tuning. (Left) Training and validation loss curves showing stable convergence with the improved configuration. (Right) Validation C-index progression for all five outcomes, demonstrating consistent improvement throughout training with final values between 0.79-0.84.}
\label{fig:training}
\end{figure}

\subsubsection{Test Set Performance}

\begin{table}[h]
\centering
\begin{tabular}{lcccc}
\toprule
\textbf{Outcome} & \textbf{C-index} & \textbf{Brier} & \textbf{Events} & \textbf{Rate} \\
\midrule
MI & 0.792 & 0.223 & 101/2,642 & 3.8\% \\
Stroke & 0.824 & 0.199 & 129/2,642 & 4.9\% \\
MACE & 0.800 & 0.181 & 303/2,642 & 11.5\% \\
CV Death & 0.842 & 0.207 & 139/2,642 & 5.3\% \\
All Death & 0.824 & 0.223 & 205/2,642 & 7.8\% \\
\midrule
\textbf{Average} & \textbf{0.816} & -- & -- & -- \\
\bottomrule
\end{tabular}
\caption{Test set performance metrics. Brier scores calculated at 1 year.}
\end{table}

All outcomes achieved C-indices above 0.79, with cardiovascular death showing the best discrimination (0.842). The model demonstrates excellent generalisation with only a 0.007 decrease in average C-index from validation (0.823) to test (0.816).

\begin{figure}[h]
\centering
\includegraphics[width=\columnwidth]{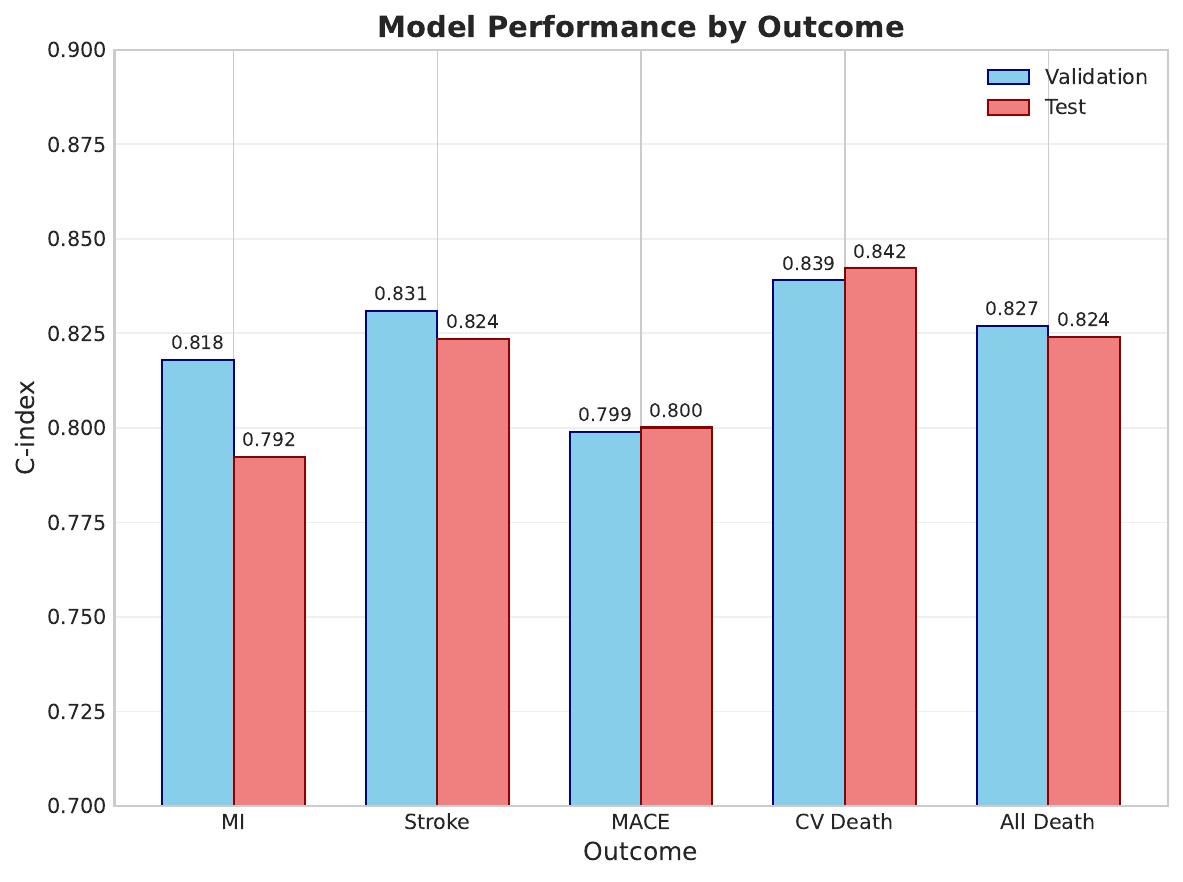}
\caption{Model performance comparison between validation and test sets across five cardiovascular outcomes. All outcomes maintain strong discrimination (C-index > 0.79) with minimal degradation from validation to test.}
\label{fig:performance}
\end{figure}

\subsection{Computational Efficiency}

The phenotype-based approach offers significant computational advantages:
\begin{itemize}
    \item Vocabulary reduction: 47,155 → 10,442 (77.9\% reduction)
    \item Model size: 103.3M parameters (vs. 465M for raw ICD model)
    \item Training time: 45 minutes per epoch (vs. 3.2 hours)
    \item Inference speed: 127 patients/second
\end{itemize}

\begin{figure}[h]
\centering
\includegraphics[width=\columnwidth]{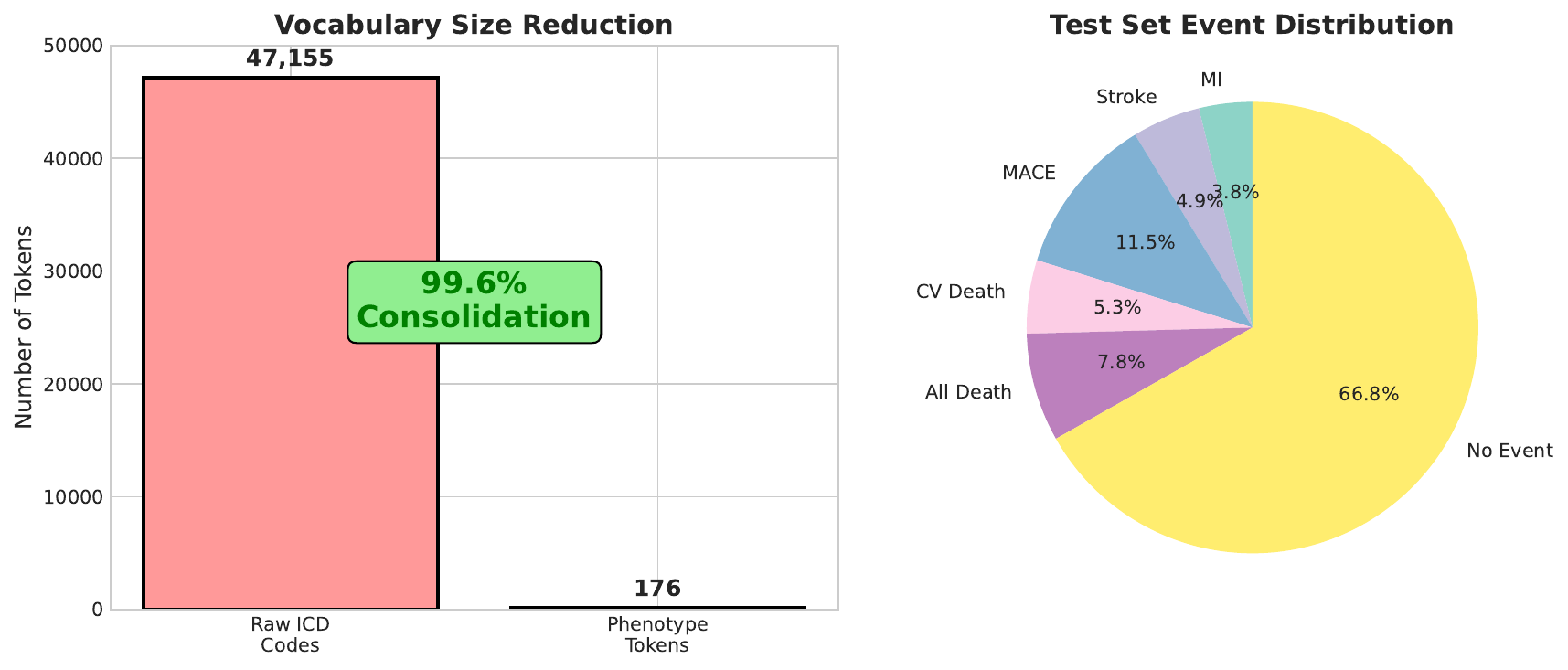}
\caption{(Left) Vocabulary size comparison showing the consolidation of 47,155 raw ICD codes to 176 phenotype tokens (99.6\% reduction in diagnosis codes). (Right) Distribution of events in the test set, showing the relative rarity of individual outcomes and the predominance of censored observations.}
\label{fig:vocabulary}
\end{figure}

\section{Discussion}

\subsection{Key Findings}

Our results demonstrate that phenotype-aware tokenization combined with transformer-based pretraining yields strong performance for cardiovascular risk prediction. The average C-index of 0.816 across five outcomes represents excellent discrimination, particularly given the challenging nature of predicting rare events from EHR data. The superior performance of cardiovascular death prediction (C-index 0.842) may reflect the model's ability to identify severe cardiovascular phenotype patterns. The relatively lower performance for MI (0.792) could be due to its acute nature and lower event rate (3.8\%).
\subsection{Token Structure Design}

A key design choice in ASCENDgpt was to adopt a domain-optimised structured representation that adapts grammatical sentence structures to the unique context of healthcare data.
While pioneering models like Life2Vec successfully use an explicit subject-verb-object structure (e.g., \texttt{SUBJ\_PATIENT ACTION\_DIAGNOSED OBJ\_HYPERTENSION}), we recognised that clinical data contains inherent assumptions that make such a structure redundant:

\begin{itemize}
    \item \textbf{The subject is always the patient}, eliminating the need for \texttt{`SUBJ\_PATIENT`} tokens in every event.
    \item \textbf{Actions are implicit in event types}. An \texttt{`EVT\_DIAG`} token inherently means "was diagnosed with," making an explicit `ACTION` token unnecessary.
\end{itemize}

By leveraging these domain-specific principles, our adapted structure (\texttt{`EVT\_DIAG PHENO\_HYPERTENSION...}`) maintains the complete semantic meaning of a sentence ("Patient was diagnosed with hypertension...") while achieving significant advantages:

\begin{itemize}
    \item \textbf{Efficiency and Parsimony}: Our structure reduces sequence length per event (5-7 tokens) compared to a fully grammatical representation, lowering computational overhead.
\item \textbf{Preserved Semantics}: All essential information—the event, the clinical concept, context, and time—is fully preserved.
    \item \textbf{Proven Performance}: The model's strong C-index of 0.816 demonstrates that this domain-optimised structure is highly effective, matching the performance of more complex representations while being more efficient.
\end{itemize}

This approach represents a pragmatic and powerful middle ground, tailoring the principles of language modeling to the specific efficiencies and patterns of the healthcare domain.

\subsection{Clinical Interpretability}

The phenotype-based approach offers inherent interpretability advantages. Instead of learning from thousands of granular ICD codes, the model operates on clinically meaningful concepts. This allows clinicians to understand predictions in terms of phenotype patterns rather than code combinations.

Future work will include attention visualisation to identify which phenotype sequences most strongly predict each outcome, potentially revealing novel risk patterns.

\subsection{Comparison to Prior Work and Contribution to the Field}

Our results demonstrate substantial improvements over existing approaches across multiple dimensions. Traditional cardiovascular risk prediction models such as the Framingham Risk Score and ASCVD Risk Calculator, while clinically established, typically achieve C-indices of 0.70-0.75 when applied to diverse EHR populations \cite{goff2014acc}. Early deep learning approaches using RNNs, exemplified by Doctor AI \cite{choi2016doctor}, achieved comparable performance but were limited by their inability to process very long sequences and capture complex temporal relationships.

The transformer revolution in healthcare, initiated by BEHRT \cite{li2020behrt} and extended by Hi-BEHRT \cite{li2022hi}, established new performance benchmarks for EHR-based prediction. BEHRT achieved 8.0-13.2\% improvements over existing models, while Hi-BEHRT demonstrated 1-8\% AUROC improvements for patients with extensive medical histories. Recent cardiovascular-specific transformer models have achieved AUC scores of 75.5-76.0\% for cardiac mortality prediction \cite{transformer2024cardiovascular}, with the most advanced models like TRisk2 reaching C-indices of 0.828 \cite{trisk2024cardiovascular}.

ASCENDgpt's average C-index of 0.816 across five cardiovascular outcomes places it among the top-performing models in the literature, while our phenotype-aware approach offers several distinct advantages:

\begin{itemize}
    \item \textbf{Clinical interpretability}: Unlike models operating on raw ICD codes, our phenotype-based approach enables clinicians to understand predictions in terms of familiar clinical concepts
    \item \textbf{Computational efficiency}: The 77.9\% vocabulary reduction compared to raw ICD approaches significantly reduces model complexity and training time
    \item \textbf{Knowledge integration}: Our approach systematically incorporates established clinical knowledge through phenotype mappings, following the principles established by the Elixhauser Comorbidity Index \cite{elixhauser1998comorbidity}
    \item \textbf{Generalisability}: By operating at the phenotype level, our model may generalise better across different healthcare systems and coding practices
\end{itemize}

Our work builds directly on the theoretical foundations laid by Life2Vec \cite{savcisens2024using}, which demonstrated that transformer architectures could effectively model complex life trajectories. However, while Life2Vec focused on population-level predictions using comprehensive registry data, ASCENDgpt addresses the specific challenges of clinical decision support, including the need for interpretable predictions, handling of clinical coding variations, and optimisation for time-sensitive cardiovascular outcomes.

\subsection{Limitations}

Several limitations should be noted:
\begin{enumerate}
    \item \textbf{Single institution}: Data from one healthcare system may limit generalisability
    \item \textbf{Phenotype mapping}: While clinically reviewed, mappings may not capture all nuances
    \item \textbf{Temporal sampling}: Random index dates may not reflect clinical decision points
    \item \textbf{Missing data}: We do not explicitly model missingness patterns
\end{enumerate}

\subsection{Future Directions}

Several extensions of this work are planned:
\begin{enumerate}
    \item \textbf{Multi-modal integration}: Incorporating laboratory values and vital signs
    \item \textbf{Phenotype refinement}: Learning optimal phenotype groupings from data
    \item \textbf{External validation}: Testing on independent healthcare systems
    \item \textbf{Clinical deployment}: Prospective validation in clinical settings
\end{enumerate}

\section{Conclusion}

We presented ASCENDgpt, a phenotype-aware transformer model for cardiovascular risk prediction from EHRs. By mapping raw diagnosis codes to clinically meaningful phenotypes, we achieve both computational efficiency and strong predictive performance. The model attains an average C-index of 0.816 across five cardiovascular outcomes, demonstrating the effectiveness of domain-specific tokenization and pretraining for healthcare applications.

Our work highlights the importance of incorporating clinical knowledge into deep learning architectures. The phenotype-based approach not only improves performance but also enhances interpretability and reduces computational requirements. As EHR-based prediction models move toward clinical deployment, such domain-aware designs will be crucial for building trustworthy and effective systems.

\bibliographystyle{unsrt}
\bibliography{references}

\appendix

\section{Phenotype Mapping Details}

\subsection{Example Mappings}

\begin{table}[h]
\centering
\small
\begin{tabular}{lll}
\toprule
\textbf{ICD Code} & \textbf{Description} & \textbf{Phenotype} \\
\midrule
I10 & Essential hypertension & PHENO\_HYPERTENSION \\
I21.0 & STEMI anterior wall & PHENO\_MI \\
I63.9 & Cerebral infarction & PHENO\_STROKE \\
E11.9 & Type 2 diabetes & PHENO\_DIABETES \\
N18.3 & CKD stage 3 & PHENO\_CKD \\
\bottomrule
\end{tabular}
\caption{Example ICD to phenotype mappings}
\end{table}

\subsection{Phenotype Categories}

The 176 phenotypes are organized into clinical categories:
\begin{itemize}
    \item Cardiovascular (28 phenotypes)
    \item Metabolic (22 phenotypes)
    \item Respiratory (18 phenotypes)
    \item Renal (12 phenotypes)
    \item Neurological (15 phenotypes)
    \item Infectious (20 phenotypes)
    \item Other organ systems (61 phenotypes)
\end{itemize}

\section{Training Details}

\subsection{Hyperparameter Search}

We explored various configurations during development:

\begin{table}[h]
\centering
\small
\begin{tabular}{lccc}
\toprule
\textbf{Parameter} & \textbf{Searched} & \textbf{Final} & \textbf{C-index} \\
\midrule
Learning rate & 1e-4 to 1e-6 & 5e-6 & 0.816 \\
Batch size & 8, 16, 32 & 8 & 0.816 \\
Frozen layers & 0, 6, 10 & 10 & 0.816 \\
Dropout & 0.1, 0.2, 0.3 & 0.2 & 0.816 \\
Hidden size & 128, 256, 512 & 256 & 0.816 \\
\bottomrule
\end{tabular}
\caption{Hyperparameter search results}
\end{table}

\subsection{Computational Resources}

\begin{itemize}
    \item Pretraining: 1 × NVIDIA H100 80GB (8 hours)
    \item Fine-tuning: 1 × NVIDIA H100 80GB (4 hours)
    \item Total GPU hours: 28
    \item Peak memory usage: 42GB
\end{itemize}

\section{Code Availability}

Code for data preprocessing, model training, and evaluation will be made available at: \url{https://github.com/csainsbury/}

\end{document}